\documentclass{article} 
\usepackage{iclr2023_workshop,times}

\usepackage{amsmath,amsfonts,bm}

\def\figref#1{figure~\ref{#1}}

\def\eqref#1{equation~\ref{#1}}

\def\1{\bm{1}}

\DeclareMathAlphabet{\mathsfit}{\encodingdefault}{\sfdefault}{m}{sl}
\SetMathAlphabet{\mathsfit}{bold}{\encodingdefault}{\sfdefault}{bx}{n}

\def\sR{{\mathbb{R}}}

\newcommand{\normltwo}{L^2}

\usepackage{hyperref}
\usepackage{url}
\usepackage{amsmath,amsfonts,bm}
\usepackage{amssymb}
\usepackage{mathtools}
\usepackage{nicefrac}       

\usepackage[newfloat, frozencache=true, cachedir=_minted_output]{minted} 
\usepackage{caption}

\usepackage[ruled,vlined]{algorithm2e}
\usepackage{tcolorbox}
\tcbuselibrary{minted,breakable,xparse,skins}

\definecolor{bg}{gray}{0.95}
\DeclareTCBListing{mintedbox}{O{}mO{}}{%
	breakable=true,
	listing engine=minted,
	listing only,
	minted language=#2,
	minted style=default,
	minted options={%
		linenos,
		gobble=0,
		breaklines=true,
		breakafter=,,
		fontsize=\small,
		escapeinside=||,
		numbersep=8pt,
		tabsize=2,
		encoding=utf8,
		#1},
	boxsep=0pt,
	left skip=0pt,
	right skip=0pt,
	left=25pt,
	right=0pt,
	top=3pt,
	bottom=3pt,
	arc=5pt,
	leftrule=0pt,
	rightrule=0pt,
	bottomrule=2pt,
	toprule=2pt,
	colback=bg,
	colframe=orange!70,
	enhanced,
	overlay={%
		\begin{tcbclipinterior}
			\fill[orange!20!white] (frame.south west) rectangle ([xshift=20pt]frame.north west);
	\end{tcbclipinterior}},
	#3}

\title{Physics-driven machine learning models coupling PyTorch and Firedrake}

\author{Nacime Bouziani \& David A. Ham
\\
Department of Mathematics \\
Imperial College London, London, UK \\
\texttt{\{n.bouziani18, david.ham\}@imperial.ac.uk} \\
}

\iclrfinalcopy 
\begin{document}

\maketitle

\begin{abstract}
Partial differential equations (PDEs) are central to describing and modelling complex physical systems that arise in many disciplines across science and engineering. However, in many realistic applications PDE modelling provides an incomplete description of the physics of interest. PDE-based machine learning techniques are designed to address this limitation. In this approach, the PDE is used as an inductive bias enabling the coupled model to rely on fundamental physical laws while requiring less training data. The deployment of high-performance simulations coupling PDEs and machine learning to complex problems necessitates  the composition of capabilities provided by machine learning and PDE-based frameworks. We present a simple yet effective coupling between the machine learning framework PyTorch and the PDE system Firedrake that provides researchers, engineers and domain specialists with a high productive way of specifying coupled models while only requiring trivial changes to existing code. 
\end{abstract}

\section{Introduction}
Partial differential equations (PDEs) are ubiquitous in science and impact natural sciences and engineering, since  most of the laws that govern the dynamics of physical systems are described by partial differential equations. This modelling is most effective when physical systems closely follow the, frequently idealised, asumptions used to derive the PDE. In more complex and realistic scenarios, the error induced by these assumptions can become large. PDE-based machine learning, or physics-driven machine learning, seeks to address this limitation by creating models informed both by the PDE and by observed data. Different discretisation techniques for the PDE part of the model are available. However, among these, the finite element method (FEM) uniquely combines a high degree of geometric and numerical flexibility with a high-level mathematical abstraction which is amenable to a differentiable programming approach. In many cases, the PDE is solved using the finite element method (FEM), a widely used method for solving PDEs, see \citep{berg_neural_2021, nguyen_tnet_2022, costabal_delta-pinns_2023}.\newline

The use of efficient, composable, and high productivity environments relying on high-level domain specific languages, and code generation is prevalent in machine learning. On the other hand, only few scientific computing tools embrace this design approach. Examples include: PyTorch \citep{paszke_pytorch_2019}, TensorFlow, \citep{abadi_tensorflow_2016}, and JAX \citep{jax2018github} for machine learning, and Firedrake \citep{Firedrake_2017}, FEniCS \citep{Fenics_2012}, or FreeFem++ \citep{FreeFem++_2012} for finite element methods (FEM), to name but a few. In this work, we extend the high productivity and high performance capabilities of PyTorch and Firedrake to physics-driven machine learning models, enabling scientists and engineers to design, implement and run complex simulations coupling machine learning models implemented in PyTorch and partial differential equations implemented in Firedrake.


\section{Firedrake}

Firedrake \citep{Firedrake_2017} is an automated finite element system embedded in Python for the solution of partial differential equations. Firedrake uses the Unified Form Language (UFL) \citep{UFL_2014}, a domain-specific language to provide high-level representations of finite element problems. The Firedrake system translates the symbolic specification of variational forms of PDEs expressed in UFL into low-level code for assembling the sparse matrices and vectors of the corresponding finite element problem. The dolfin-adjoint automatic differentiation package \citep{farrell_automated_2013, dolfin-adjoint_2019} provides Firedrake with automatically derived tangent-linear and adjoint capabilities analogous to the forward- and back-propagated derivatives available in PyTorch.


\section{Differentiable programming}

In order to run hybrid models coupling partial differential equations and machine learning, we need to be able to couple the evaluation of both components but also to couple their differentiation as gradient calculation is critical in both machine learning and PDEs. Examples include the use of backpropagation for training neural networks in machine learning, or the use of Newton-type methods for solving PDEs. The use of automatic differentiation to automate the calculation of the derivatives of interest is widely employed in machine learning while only adopted by few PDE-based software. More specifically, there are two approaches considered: tangent-linear mode, also referred to as forward mode, and the adjoint mode, also referred to as reverse mode, of which backpropagation is a special case.\newline


A central idea in differentiable programming is that the implementation of a given function $f$ as a computer program can be represented as a directed acyclic graph whose nodes and edges represent the intermediate calculations to compute $f$. Another important idea which follows from chain rule is that the computation of a derivative in tangent-linear mode (resp. adjoint-mode) of $f$ can be achieved by traversing that computational graph forward (resp. backward) and evaluate the derivative tangent linear model (resp. adjoint model) of each node on the fly. \newline

For example, let $V$ be a Hilbert space and let $u$ be the solution of a PDE defined by:
\begin{equation}
\label{eq:variational_form}
F(u, m; v) = 0\quad \forall v \in V    
\end{equation}
where $F$ is the variational form of the PDE, and $m$ is a known parameter. Then, backpropagating through the solution $u$ is equivalent to compute the adjoint model of $u$, which can be written as:
\begin{equation}
    \label{eq:TLM_PDE}
    \mathcal{J}^{*}_{u}(m; w) = - \frac{\partial F}{\partial m}^{*} \lambda
\end{equation}
for all $w \in V^{*}$ and where $\lambda \in V$ is the solution of the adjoint equation defined as:
\begin{equation}
    \label{eq:Adjoint_equation}
    \frac{\partial F}{\partial u}^{*} \lambda = w
\end{equation}


See appendix \ref{sec:adjoint_method} for more details. Firedrake and PyTorch are both equipped with AD modules to calculate the tangent-linear and adjoint models of a wide range of operations defined within both frameworks.

\section{Coupling PyTorch and Firedrake}

Let $\mathcal{F}$ be an operator representing a set of operations implemented in Firedrake (e.g. solving a PDE, or assembling a variational form). The use of the Firedrake finite element system for building physics-driven machine learning models within PyTorch necessitates the ability for PyTorch to go through the computational graph of $\mathcal{F}$ to evaluate it and automatically differentiate it. Note that the other side of the coupling, namely using PyTorch operations within Firedrake is already possible as introduced in \citet{bouziani_ham_2021}. Also, similar work is conducted to compose Firedrake with JAX programs in \citet{yashchuk_bringing_2022}.\newline

\begin{figure}[ht!]
\centering
\includegraphics[width=.70\linewidth]{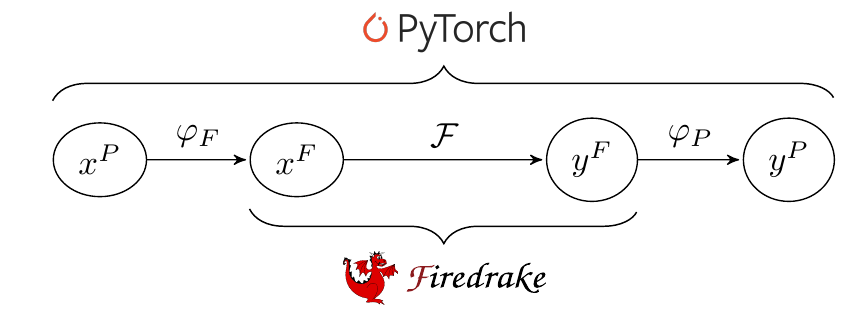}
\caption{Subgraph of the PyTorch computational graph containing Firedrake operations of interest represented by $\mathcal{F}$, where $P$ refers to PyTorch variables and $F$ to Firedrake variables. $\varphi_{F}$ and $\varphi_{P}$ represent the casting of a PyTorch tensor to a Firedrake Function and vice versa.}
\label{fig:coupling_firedrake_pytorch}
\end{figure}

 In this work, we build a custom Firedrake operator within \emph{torch.autograd} \citep{paszke_automatic_2017} to represent $\mathcal{F}$. The computational graph evaluation and differentiation is delegated to \emph{torch.autograd} for PyTorch-based nodes, and to \emph{dolfin-adjoint} for Firedrake-based nodes. This simple yet powerful high-level coupling, illustrated in \figref{fig:coupling_firedrake_pytorch}, results in a composable environment that benefits from the full armoury of advanced features and AD capabilities both frameworks offer whilst maintaining separation of concerns.\newline

This coupling is facilitated by the fact that the core data object in each case is merely a tensor of values. However, machine learning frameworks do not inherently attach further structure to this tensor. In contrast, FEM packages such as Firedrake explicitly associate state tensors with computational domain information. Consequently, the definition of Firedrake tensors induces the required mapping $\varphi_{F}$ and $\varphi_{P}$ to convert to and from PyTorch, see appendix \ref{sec:custom_mappings} for more details. Another consequence is that the Firedrake tensor representation facilitates the use of appropriate inner products, such as $L^p$ and $H^p$,  in addition to the $\ell^{p}$-norm ubiquitous in machine learning. This is critical to guarantee mesh independence for PDE-constrained optimisation problems, see \citep{schwedes_mesh_2017}.\newline

Listing \ref{code:coupling_template} demonstrates how one can build physics-driven models using PyTorch and Firedrake. The fact that both libraries are embedded in Python facilitates the implementation of such models. More importantly, note that coupling both frameworks only requires a one-line change to existing codes as highlighted in listing \ref{code:coupling_template} (cf.line 12), which makes the implementation straightforward from a user perspective. Line 9 defines the functional $\mathcal{F}$ as a function of given control(s), which enables to only traverse the relevant part of $\mathcal{F}$'s computational graph needed to differentiate $\mathcal{F}$ with respect to the given control(s).  We also extend dolfin-adjoint to relax the assumption that it owns the final quantity of interest that is differentiated, thereby enabling backpropagation computation to start from a PyTorch tensor. The operator $G$ defined in line 12 wraps the operator $F$ to act on PyTorch tensors as illustrated in line 15 with $x^{P}$ and $y^{P}$.
\begin{listing}[ht!]
\captionof{listing}{Outline of backpropagation through Firedrake using $\operatorname{torch\_operator}$}
\label{code:coupling_template}
\begin{mintedbox}[highlightlines={12}]{python}
import torch
import firedrake as fd
import firedrake.ml as fd_ml
import firedrake_adjoint as fda

...

# Defined reduced functional |$\mathcal{F}$| with respect to control(s)
F = fda.|\textcolor{blue}{ReducedFunctional}|(y_F, |\textcolor{blue}{Control}|(x_F))

# Define the coupling operator: |$G \vcentcolon= \varphi_{P} \circ \mathcal{F} \circ \varphi_{F}$|
G = fd_ml.|\textcolor{blue}{torch\_operator}|(F)

# Apply the coupling operator to a torch.Tensor |$x^{P}$|
y_P = G(x_P)

# Backpropagate through |$G$|: calculate |$\mathcal{J}^{*}_{y^{P}}(x^{P}, w^{P})$|
w_P = ...
y_P.backward(w_P)
\end{mintedbox}
\end{listing}

\section{Example: 2D heat conductivity inverse problem}

We showcase our implementation on a simple heat conductivity problem modelled via the following time-independent heat equation: Find the temperature field $u \in V$ such that:
\begin{equation}
    \label{eq:heat_conductivity_strong_form}
    \begin{aligned}
        - \nabla \cdot \left( e^{\kappa} \nabla u\right) &= f \quad \text{in } \Omega\\
        u &= 0 \quad \text{on } \partial \Omega
    \end{aligned}
\end{equation}
where $\Omega \subset \sR^{2}$ is an open and bounded domain, $\kappa \in V$ is the conductivity, $f \in V$ the source term, and with $V$ a suitable function space (i.e. $V = H^{1}_{0}(\Omega)$). We want to solve the inverse problem driven by \eqref{eq:heat_conductivity_strong_form} by learning the inverse operator:
\begin{equation}
    \label{eq:inverse_operator}
    \hat{\kappa}: u^{obs} \rightarrow \kappa
\end{equation}
where $u^{obs}$ refers to an observed temperature field and is modelled by: $u^{obs} = u(\kappa) + \varepsilon$, with $u(\kappa)$ the solution of \eqref{eq:heat_conductivity_strong_form} for a given $\kappa$, and $\varepsilon$ the observation noise. We use a model-constrained deep neural network approach as introduced in \citet{nguyen_tnet_2022}. More precisely, let $\kappa_{\theta}$ be a a neural network based model with $\theta$ the model parameters. We train $\kappa_{\theta}$ to learn the inverse operator defined in \eqref{eq:inverse_operator} by minimising the following loss function:
\begin{equation}
    \label{eq:tnet_J}
    J(u^{obs}, \kappa^{exact}; \theta) = \frac{1}{2}\|\kappa_{\theta}(u^{obs}) - \kappa^{exact}\|^{2}_{\normltwo(\Omega)} + \frac{\alpha}{2} \|u(\kappa_{\theta}(u^{obs})) - u^{obs}\|_{\normltwo(\Omega)}^{2}
\end{equation}
where $\kappa^{exact}$ refers to the exact conductivity for a given observable $u^{obs}$. In \citet{nguyen_tnet_2022}, authors studied heat conductivity test case using a single hidden layer model, here we consider a convolutional neural network architecture (CNN) for learning the conductivity. Two Firedrake coupling operators are used during training: one for computing the PDE solution $u(\cdot)$, and the other for assembling the $L^2$-loss as the choice of norm is critical for mesh-independence. For training, we generate $n$ random fields to form the training split: $\{\kappa^{exact}_{i}, u^{obs}_{i}\}_{1 \le i \le n}$, and average the loss defined in \eqref{eq:tnet_J} across training samples. For evaluation, we generate data in the same way and average the error across test samples. More details about this test case can be found in appendix \ref{sec:implementation_details}. The exact version of Firedrake used in this paper is archived in \citet{zenodo/Firedrake-20230208.0} while the code used to run this example is archived in \citet{bouziani_physics-driven_2023}.
\begin{figure}[htp]
\centering
\includegraphics[width=.45\linewidth]{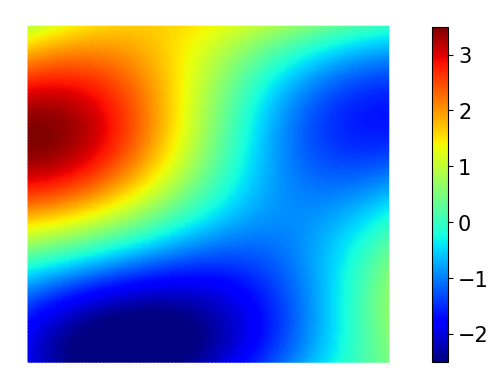}
\includegraphics[width=.45\linewidth]{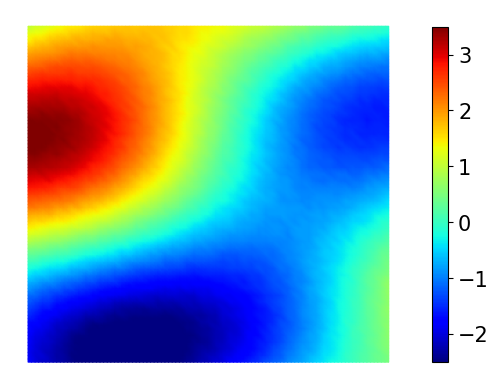}
\caption{Heat conductivity as a function of position $(x, y)$: exact conductivity (left), conductivity reconstructed from observed data $\kappa_{\theta}(u^{obs})$ by the CNN model (right).}
\label{fig:heat_conductivity_models}
\end{figure}
\subsubsection*{Acknowledgments}
This work was funded by a President’s PhD Scholarship at Imperial College London.

\bibliography{iclr2023_workshop}
\bibliographystyle{iclr2023_workshop}

\newpage
\appendix
\section{Appendix}

\subsection{Adjoint model of PDE solutions}
\label{sec:adjoint_method}

Let $V$ and $M$ be Hilbert spaces, and let $u \in V$ be the solution of the parametrised partial differential equation defined as:
\begin{equation}
    \label{eq:appendix:adjoint:F_residual}
    F(u, m;  v) = 0 \quad \forall v \in V
\end{equation}

with $F$ the variational form of the PDE, and $m \in M$ the control value. We assume that the PDE defined by $F$ yields a unique solution for any control value $m \in M$. Hence, we can define the PDE solution operator $u(\cdot) \vcentcolon M \rightarrow V$ that maps any control value $m$ to the corresponding solution $u(m)$ satisfying \eqref{eq:appendix:adjoint:F_residual}. Note that the PDE solution $u(m)$ is generally not known explicitly but can be approximated using the finite element method. \\

We assume that the linear form $F$ is continuously Fr\'echet differentiable and that the linearised PDE operator $\frac{\partial F}{\partial u}$ is invertible. It follows, using implicit function theorem, that the solution operator $u(\cdot)$ is continuously Fr\'echet differentiable, see \citep[Sec. 1.4.2]{hinze_optimization_2009}. Let $V^{*}$ and $M^{*}$ be the dual spaces of $V$ and $M$, i.e. the spaces of all continuous linear functionals on $V$ and $M$, respectively. Backpropagating through the PDE solution $u(m)$ equates to computing its adjoint model $\mathcal{J}^{*}_{u} \vcentcolon V^{*} \rightarrow M^{*}$ defined as:
\begin{equation}
    \label{eq:appendix:adjoint:adj_model_def}
    \mathcal{J}^{*}_{u}(m; w) = \frac{du}{dm}^{*} w \quad \forall w \in V^{*}
\end{equation}

where $\frac{du}{dm}^{*} \in \mathcal{L}(V^{*}, M^{*})$ is the adjoint of the G\^ateaux derivative of $u$ with respect to $m$. While the expression of $\frac{du}{dm}^{*}$ is not known explicitly, given that $u(m)$ is the PDE solution, we can still compute it using \eqref{eq:appendix:adjoint:F_residual}. More precisely, given that we have $F(u(m), m; v) = 0$ for all $m \in M$, it follows that the derivative $\frac{dF}{dm}$ must be zero. Therefore, applying chain rule yields:
\begin{equation}
    \label{eq:appendix:adjoint:dFdm_0}
    \frac{\partial F}{\partial u} \frac{du}{dm} + \frac{\partial F}{\partial m} = 0
\end{equation}

which leads to
\begin{equation}
    \label{eq:appendix:adjoint:dudm}
    \frac{du}{dm} = - \frac{\partial F}{\partial u}^{-1} \frac{\partial F}{\partial m}
\end{equation}

since we assumed $\frac{\partial F}{\partial u}$ to be invertible. Combining equations \ref{eq:appendix:adjoint:dudm} and \ref{eq:appendix:adjoint:adj_model_def}, we finally obtain for all $w \in V^{*}$:
\begin{equation}
    \label{eq:appendix:adjoint:adj_model}
    \mathcal{J}^{*}_{u}(m; w) = - \frac{\partial F}{\partial m}^{*} \lambda
\end{equation}

where $\lambda \in V$ is the solution of the adjoint equation defined as
\begin{equation}
    \label{eq:appendix:adjoint:adj_equation}
    \frac{\partial F}{\partial u}^{*} \lambda = w
\end{equation}

We refer the interested reader to \citep{schwedes_mesh_2017} and \citep[Sec. 1.6]{hinze_optimization_2009} for more details. 

\subsection{Custom mappings}
\label{sec:custom_mappings}

Depending on the user case, different representations can be adopted to represent finite element functions in PyTorch. For example, one can feed a neural network with the values of a finite element function on a set of points, e.g on a uniform Cartesian grid for CNN-based architectures or a more general grid for graph neural networks. This representation is independent of the mappings $\varphi_{F}$ and $\varphi_{P}$ illustrated in \figref{fig:coupling_firedrake_pytorch} which merely casts tensors from one framework to the other without altering their shape. This consideration is rather an operation that happens either in Firedrake and/or PyTorch. \newline

In particular, complex representations of finite element functions can be lifted at the level of the definition of the mesh and function spaces in Firedrake, as it inherently provides a richer space representation than PyTorch.

\subsection{2D heat conductivity inverse problem: implementation details}
\label{sec:implementation_details}

We generate a synthetic dataset $\{\kappa^{exact}_{i}, u^{obs}_{i}\}_{1 \le i \le n}$ for training and test using the following procedure:

\begin{itemize}
    \item Randomly generate parameter of interest $\{\kappa_{i}\}_{1 \le i \le n}$
    \item Compute for each $\kappa_{i}$ the corresponding solution of the forward problem (PDE): $u(\kappa_{i})$
    \item Add noise to form the observables: $u_{i}^{obs} = u(\kappa_{i}) + \varepsilon_{i} \quad\forall i \in [|1, n|]$, e.g. $\varepsilon_{i} \in \mathcal{N}(0, 1)$.
\end{itemize}

This procedure is not specific to a given PDE and can be adapted to a wide range of inverse problems. For the heat conductivity example, we generate 500 training samples and 100 test samples. For training, we average the loss defined in \eqref{eq:tnet_J} over the $n$ training samples, i.e. we have:

\begin{equation}
    \label{eq:total_loss}
    \mathcal{L} = \frac{1}{2n} \sum_{i=1}^{n} \left( \|\kappa_{\theta}(u^{obs}_{i}) - \kappa^{exact}_{i}\|^{2}_{\normltwo(\Omega)} + \alpha\|u(\kappa_{\theta}(u^{obs}_{i})) - u^{obs}_{i}\|_{\normltwo(\Omega)}^{2} \right)
\end{equation}

For evaluation, we adopt the evaluation metric used in  \citet{nguyen_tnet_2022}, i.e. we report the average relative error on the $m$ test samples:

\begin{equation}
    \label{eq:evaluation_metric}
    R = \frac{1}{m} \sum_{i=1}^{m} \frac{\|\kappa_{\theta}(u^{obs}_{i}) - \kappa^{exact}_{i}\|^{2}_{L^{2}(\Omega)}}{\|\kappa^{exact}_{i}\|^{2}_{L^{2}(\Omega)}}
\end{equation}

The heat conductivity example is mostly designed for demonstrating the criticality of coupling PyTorch and Firedrake to design, implement, and run complex physics-driven machine learning models in a highly productive way. Despite neither having a baseline to compare with nor an available dataset to report the performance of our trained CNN architecture, we release the dataset we generated for this example for sake of future comparisons. On this synthetic dataset, the average relative $L^2$-error $R$ on the test split is $17.68\%$. This result is not directly comparable with the analogous example presented in \citet{nguyen_tnet_2022} as the dataset and the problem formulation differ. However, given the close similarity between both test cases and the significant gap in performance in favour of our model, we argue that this can be explained by the fact that we have considered a more complex architecture. \newline

We also provide the base implementation for implementing physics-driven machine learning models using PyTorch and Firedrake for arbitrarily defined PDE-based systems in Firedrake by providing support for data generation, training, and evaluation. The corresponding package can be found in \citet{bouziani_physics-driven_2023}.

\end{document}